\documentclass[11pt,a4paper]{article}
\usepackage[hyperref]{acl2017}
\usepackage{times}
\usepackage{latexsym}
\usepackage[fleqn]{amsmath}
\usepackage{amsfonts}
\usepackage{amssymb}
\usepackage{amsthm}
\usepackage[bb=fourier]{mathalfa}
\usepackage{microtype}
\usepackage{verbatim}
\usepackage{multirow}
\usepackage{multicol}
\usepackage{graphicx}
\usepackage[footnotesize]{caption}
\usepackage{soul}
\usepackage{relsize}
\usepackage{algorithm}
\usepackage{algorithmicx}
\usepackage[noend]{algpseudocode}
\usepackage{bm}

\theoremstyle{definition}
\newtheorem{prop}{Proposition}

\newcommand{\obs}{x}
\newcommand{\msg}{z}
\newcommand{\act}{u}
\newcommand{\lobs}[2]{x_{#1}^{(#2)}}
\newcommand{\lmsg}[2]{z_{#1}^{(#2)}}
\newcommand{\lact}[2]{u_{#1}^{(#2)}}
\newcommand{\belief}{\beta}

\newcommand{\tr}{\textit{tr}}
\newcommand{\expect}{\mathbb{E}}

\newcommand{\kl}{\mathcal{D}_{\textrm{KL}}}
\DeclareMathOperator*{\argmax}{arg\,max}
\DeclareMathOperator*{\argmin}{arg\,min}

\usepackage{url}

\aclfinalcopy 

\title{Translating Neuralese}

\author{Jacob Andreas ~~ Anca Dragan ~~ Dan Klein \\
  Computer Science Division \\
  University of California, Berkeley \\
  {\tt \{jda,anca,klein\}@cs.berkeley.edu}
}

\date{}

\begin{document}
\def\arraystretch{1.2}

\maketitle

\begin{abstract}
  Several approaches have recently been proposed for learning decentralized deep
  multiagent policies that coordinate via a differentiable communication channel.
  While these policies are effective for many tasks, interpretation of
  their induced communication strategies has remained a challenge. Here we
  propose to interpret agents' messages by translating them.  
  Unlike in typical machine translation problems, we have no parallel data
  to learn from. Instead we develop a translation model based on the insight
  that agent messages and natural language strings mean the same thing \emph{if they
  induce the same belief about the world in a listener}.
  We present theoretical guarantees and empirical evidence that our approach
  preserves both the semantics and pragmatics of messages by ensuring that
  players communicating through a translation layer do not suffer a substantial
  loss in reward relative to players with a common language.\footnote{
  We have released code and data at \url{http://github.com/jacobandreas/neuralese}.
  }
\end{abstract}

\section{Introduction}
Several recent papers have described approaches for learning
\emph{deep communicating policies} (DCPs): decentralized representations of
behavior that enable multiple agents to communicate via a
differentiable channel that can be formulated as a recurrent neural network. 
DCPs have been shown to solve a variety of coordination problems,
including reference games \cite{Lazaridou16Communication}, logic puzzles
\cite{Foerster16Communication}, and simple control \cite{Sukhbaatar16CommNet}.
Appealingly, the agents' communication protocol can be learned via direct
backpropagation through the communication channel, avoiding many of the
challenging inference problems associated with learning in classical
decentralized decision processes \cite{Roth05DecComm}. 

\begin{figure}[t]
  \centering
  \vspace{-1em}
  \includegraphics[width=\columnwidth,clip,trim=0in 4.8in 5in 0in]{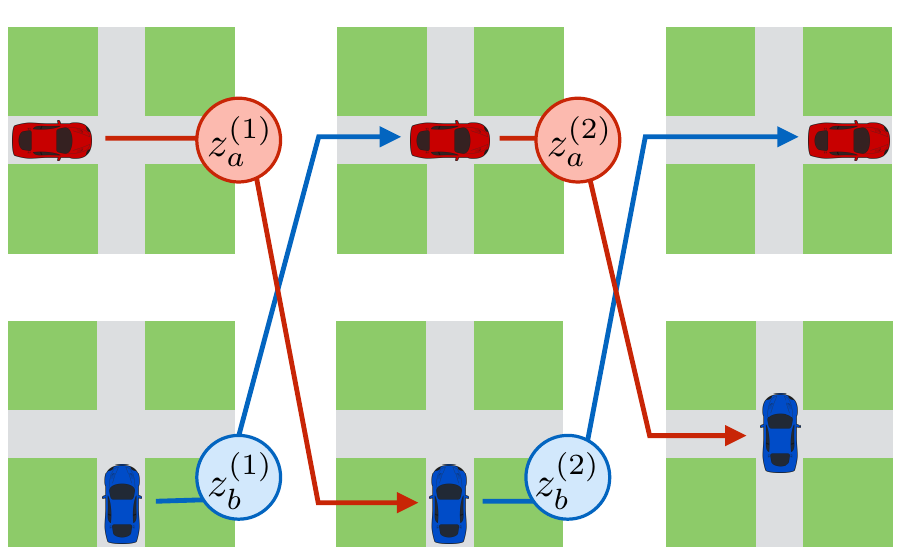}
  \vspace{-1.5em}
  \caption{
    Example interaction between a pair of agents in a deep communicating policy.
    Both cars are attempting to cross the intersection, but cannot see each
    other. By exchanging message vectors $\msg^{(t)}$, the agents are able to
    coordinate and avoid a collision. This paper presents an approach for
    \emph{understanding} the contents of these message vectors by translating
    them into natural language.
  }
  \label{fig:teaser}
  \vspace{-1em}
\end{figure}

But analysis of the strategies induced by DCPs has remained a challenge.  As an
example, \autoref{fig:teaser} depicts a driving game in which two cars, which
are unable to see each other, must both cross an intersection without colliding.
In order to ensure success, it is clear that the cars must communicate with each
other. But a number of successful communication strategies are possible---for
example, they might report their exact $(x, y)$ coordinates at every timestep,
or they might simply announce whenever they are entering and leaving the
intersection. If these messages were communicated in natural language, it would
be straightforward to determine which strategy was being employed. However, DCP agents 
instead communicate with an automatically induced protocol of unstructured, real-valued
recurrent state vectors---an artificial language we might call ``neuralese,''
which superficially bears little resemblance to natural language, and thus
frustrates attempts at direct interpretation.

We propose to understand neuralese messages by \emph{translating} them. In this
work, we present a simple technique for inducing a dictionary that maps between
neuralese message vectors and short natural language strings, given only
examples of DCP agents interacting with other agents, and humans interacting
with other humans. Natural language already provides a rich set of tools for
describing beliefs, observations, and plans---our thesis is that these tools
provide a useful complement to the visualization and ablation techniques used
in previous work on understanding complex models \cite{Strobelt16RNNVis,Ribeiro16LIME}.

While structurally quite similar to the task of machine translation between
pairs of human languages, interpretation of neuralese poses a number of novel
challenges. First, there is no natural source of parallel data: there are no bilingual
``speakers'' of both neuralese and natural language. Second, there may not be a direct
correspondence between the strategy employed by humans and DCP agents:
even if it were constrained to communicate using natural language, an automated agent
might choose to produce a different message from humans in a given state.
We tackle
both of these challenges by appealing to the grounding of messages in gameplay.
Our approach is based on one of the core insights in natural language
semantics: messages (whether in neuralese or natural language) have similar
meanings \emph{when they induce similar beliefs about the state of the world}.

\begin{figure}[t]
  \vspace{-.1em}
  \centering
  \includegraphics[width=\columnwidth,clip,trim=.1in 3.3in .1in .1in]{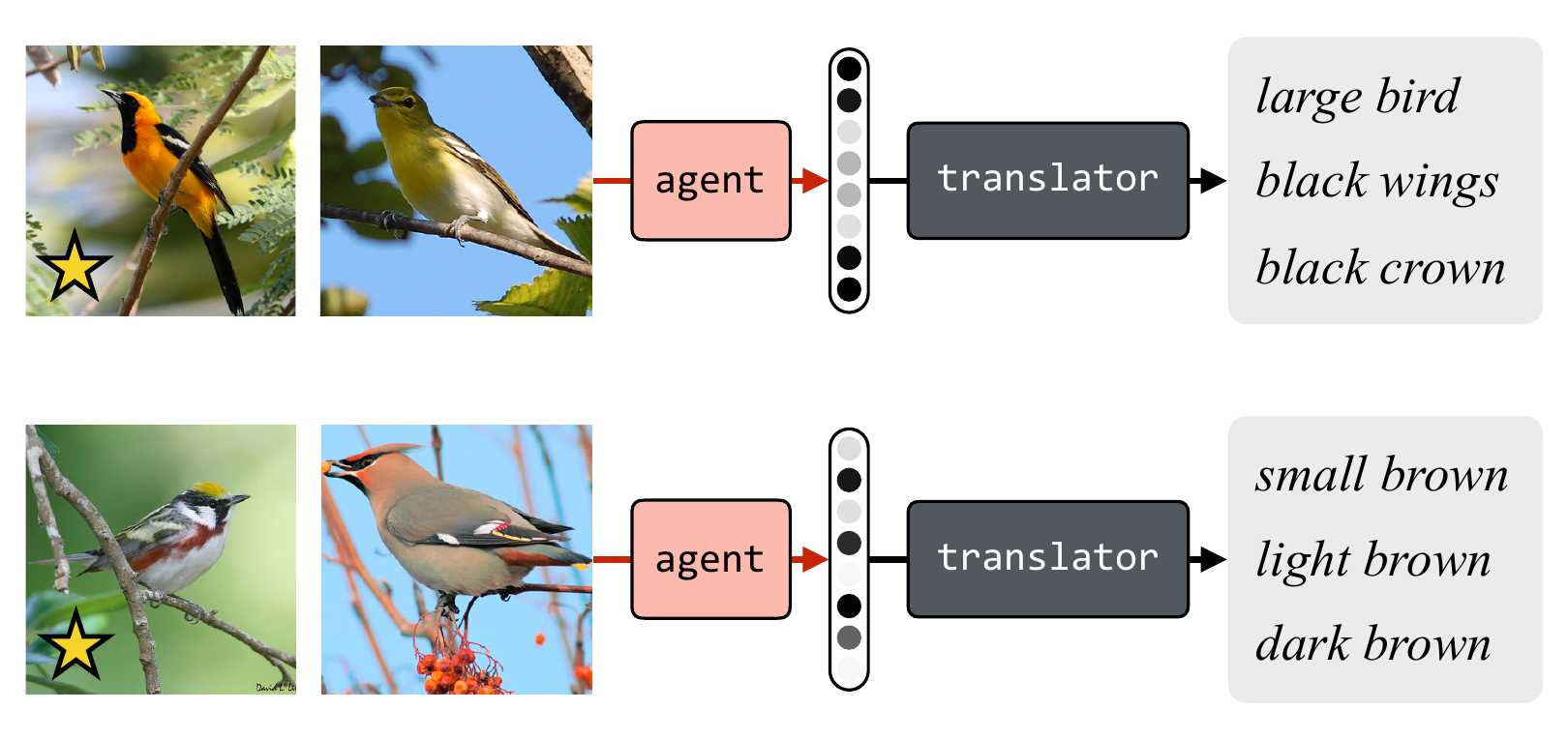}
  \vspace{-2em}
  \caption{Overview of our approach---best-scoring translations generated for a reference game involving images of birds. The speaking agent's goal is to send a message that uniquely identifies the bird on the left. From these translations
  it can be seen that the learned model appears to discriminate based on coarse attributes like size and color.}
  \label{fig:bird-examples}
\end{figure}

Based on this intuition, we introduce a translation criterion that matches
neuralese messages with natural language strings by minimizing statistical
distance in a common representation space of distributions over speaker states.
We explore several related questions:
\begin{itemize}
  \item What makes a good translation, and under what conditions is translation
    possible at all? (\autoref{sec:philosophy})
	\item How can we build a model to translate between neuralese and natural
    language? \linebreak (\autoref{sec:models})
  \item What kinds of theoretical guarantees can we provide about the behavior
    of agents communicating via this translation model? \linebreak (\autoref{sec:math})
\end{itemize}
Our translation model and analysis are general, and in fact apply equally to
human--computer and human--human translation problems grounded in gameplay. In
this paper, we focus our experiments specifically on the problem of interpreting
communication in deep policies, and apply our approach to the driving game in 
\autoref{fig:teaser} and two reference games of the kind shown in
\autoref{fig:bird-examples}. We find that this approach outperforms a more
conventional machine translation criterion both when attempting to interoperate
with neuralese speakers and when predicting their state.

\section{Related work}

A variety of approaches for learning deep policies with communication were
proposed essentially simultaneously in the past year. We have broadly labeled
these as ``deep communicating policies''; concrete examples include
\newcite{Lazaridou16Communication}, \newcite{Foerster16Communication}, and
\newcite{Sukhbaatar16CommNet}. The policy representation we employ in this paper
is similar to the latter two of these, although the general framework is
agnostic to low-level modeling details and could be straightforwardly applied to
other architectures.  Analysis of communication strategies in all these papers has
been largely ad-hoc, obtained by clustering states from which similar messages
are emitted and attempting to manually assign semantics to these clusters. The
present work aims at developing tools for performing this analysis
automatically.

Most closely related to our approach is that of
\newcite{Lazaridou16LanguageGame}, who also develop a model for assigning
natural language interpretations to learned messages;
however, this approach relies on supervised cluster labels and
is targeted specifically towards referring expression games. Here we attempt to
develop an approach that can handle general multiagent interactions without
assuming a prior discrete structure in space of observations.

The literature on learning decentralized multi-agent policies in general is
considerably larger \cite{Bernstein02DecPOMDP, Dibangoye16DecPOMDP}.
This includes work focused on communication in multiagent settings
\cite{Roth05DecComm} and even communication using natural language messages
\cite{Vogel13DecPOMDP}. All of these approaches employ structured
communication schemes with manually engineered messaging protocols; these are,
in some sense, automatically interpretable, but at the cost of introducing
considerable complexity into both training and inference.

Our evaluation in this paper investigates communication strategies that arise
in a number of different games, including reference games and an
extended-horizon driving game. Communication strategies for reference games
were previously explored by \newcite{Vogel13Grice},
\newcite{Andreas16Pragmatics} and \newcite{Kazemzadeh14ReferIt}, and reference
games specifically featuring end-to-end communication protocols by
\newcite{Yu16Reinforcer}. On the control side, a long line of work considers
nonverbal communication strategies in multiagent policies
\cite{Dragan13Legibility}. 

Another group of related approaches focuses on the development of more general
machinery for interpreting deep models in which messages have no explicit
semantics. This includes both visualization techniques
\cite{Zeiler14Vis,Strobelt16RNNVis}, and approaches focused on
generating explanations in the form of natural language
\cite{Hendricks16Explanations,Vedantam17RefExp}.

\section{Problem formulation}
\label{sec:formulation}

\paragraph{Games}

\begin{figure}
  \vspace{-.5em}
  \centering
  \includegraphics[width=0.8\columnwidth,clip,trim=0in 1.7in 5in 0in]{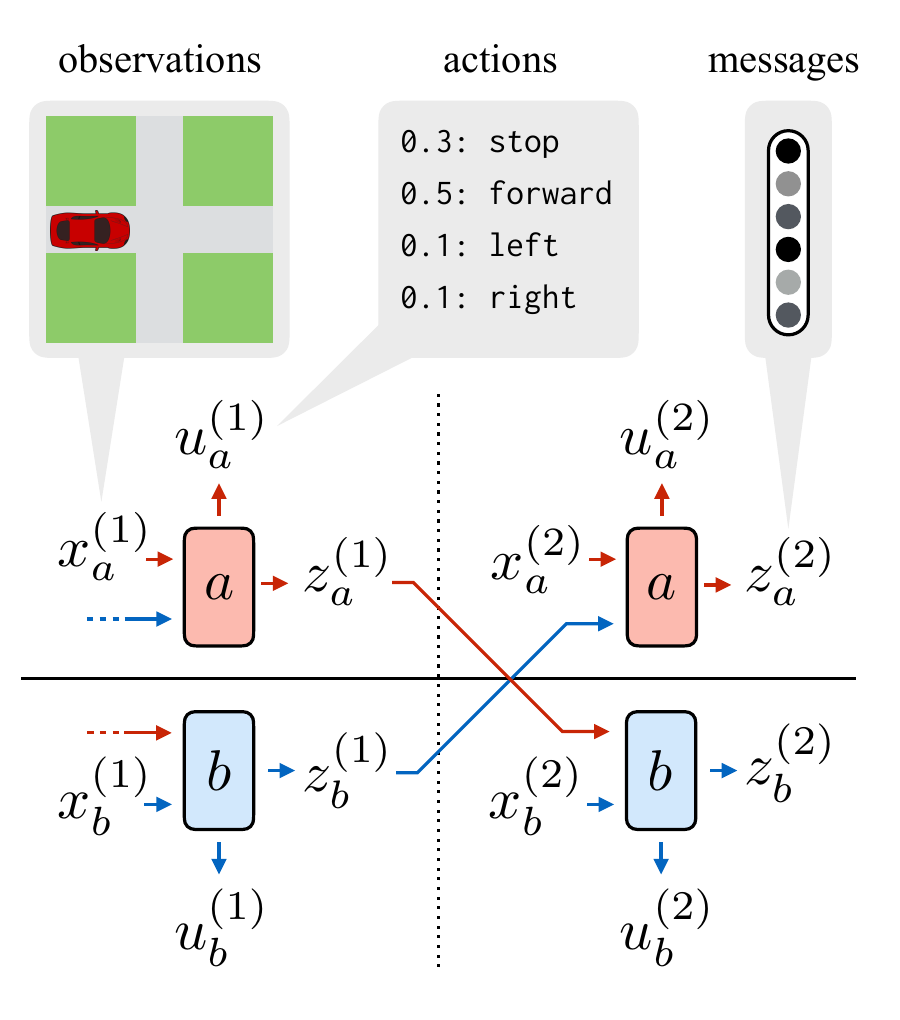} \\
  \vspace{-1em}
  \caption{
    Schematic representation of communication games. At every timestep $t$,
    players $a$ and $b$ make an observation $\obs^{(t)}$ and receive a message
    $\msg^{(t-1)}$, then produce an action $\act^{(t)}$ and a new message
    $\msg^{(t)}$. 
  }
	\label{fig:model}
  \vspace{-1em}
\end{figure}

Consider a cooperative game with two players $a$ and $b$ of the form given in
\autoref{fig:model}. At every step $t$ of this game, player $a$ makes an observation
$\lobs{a}{t}$ and receives a message $\lmsg{b}{t-1}$ from $b$. It then takes an
action $\lact{a}{t}$ and sends a message $\lmsg{a}{t}$ to $b$. (The process is
symmetric for $b$.) The distributions $p(\act_a|\obs_a,\msg_b)$ and
$p(\msg_a|\obs_a)$
together define a policy $\pi$ which we assume is shared by both players, i.e.\
$p(\act_a|\obs_a,\msg_b) = p(\act_b|\obs_b,\msg_a)$ and $p(\msg_a|\obs_a) =
p(\msg_b|\obs_b)$.  As in a standard Markov decision process, the actions
$(\lact{a}{t}, \lact{b}{t})$ alter the world state, generating new
observations for both players and a reward shared by both.

The distributions $p(\msg|\obs)$ and $p(\act|\obs,\msg)$ may also be viewed
as defining a \emph{language}: they specify how a speaker will generate messages based on
world states, and how a listener will respond to these messages.  Our goal in
this work is to learn to translate between pairs of languages generated by
different policies.  Specifically, we assume that we have access to two
policies for the same game: a ``robot policy'' $\pi_r$ and a ``human
policy'' $\pi_h$. We would like to use the representation of $\pi_h$, the
behavior of which is transparent to human users, in order to \emph{understand}
the behavior of $\pi_r$ (which is in general an uninterpretable learned model);
we will do this by inducing bilingual dictionaries that map message vectors $\msg_r$
of $\pi_r$ to natural language strings $\msg_h$ of $\pi_h$ and vice-versa.

\paragraph{Learned agents $\bm{\pi_r}$}

\begin{figure}[b!]
\vspace{-1.5em}
  \centering
  \includegraphics[width=0.75\columnwidth,clip,trim=0in 5in 4.7in 0.1in]{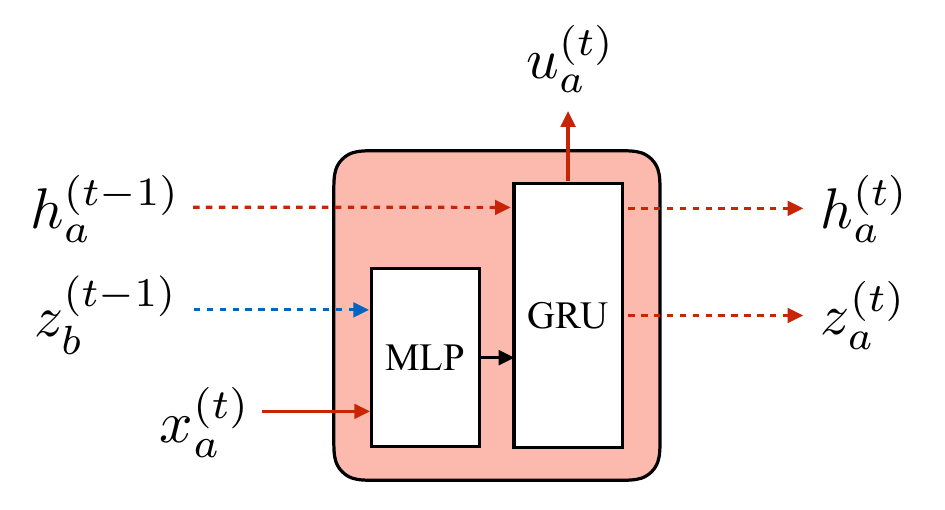}
  \caption{Cell implementing a single step of agent communication (compare with
    \newcite{Sukhbaatar16CommNet} and \newcite{Foerster16Communication}).
    \emph{MLP} denotes a multilayer perceptron; \emph{GRU} denotes a gated
    recurrent unit \cite{Cho14GRU}. Dashed lines represent recurrent
    connections.}
  \label{fig:agent}
  \vspace{-.2em}
\end{figure}

Our goal is to present tools for interpretation of learned messages that are
agnostic to the details of the underlying algorithm for acquiring them. We use a
generic DCP model as a basis for the techniques developed in this paper.  Here
each agent policy is represented as a deep recurrent Q network
\cite{Hausknecht15DRQN}. This network is built from communicating cells of the
kind depicted in \autoref{fig:agent}.  At every timestep, this agent receives
three pieces of information: an observation of the current state of the
world, the agent's memory vector from the previous timestep, and a message from
the other player. It then produces three outputs: a predicted Q value for every
possible action, a new memory vector for the next timestep, and a message to
send to the other agent.

\newcite{Sukhbaatar16CommNet} observe that models of this form may be viewed as
specifying a single RNN in which weight matrices have a particular block
structure. Such models may thus be trained using the standard recurrent
Q-learning objective, with communication protocol learned
end-to-end.

\paragraph{Human agents $\bm{\pi_h}$} The translation model we develop requires a 
representation of the distribution over messages $p(z_a|x_a)$ employed by human speakers
(without assuming that humans and agents produce equivalent messages
in equivalent contexts). We model the human message generation process as categorical,
and fit a simple multilayer perceptron model to map from observations to words and 
phrases  used during human gameplay.

\section{What's in a translation?}
\label{sec:philosophy}

What does it mean for a message $\msg_h$ to be a ``translation'' of a message
$\msg_r$? In standard machine translation problems, the answer is that $\msg_h$
is likely to co-occur in parallel data with $\msg_r$; that is, $p(\msg_h |
\msg_r)$ is large. 
Here we have no parallel data: even if we could observe natural language and neuralese
messages produced by agents in the same state, we would have no guarantee that these
messages actually served the same function.
Our answer must instead
appeal to the fact that both natural language and neuralese messages are
grounded in a common environment. For a given neuralese message $\msg_r$, we
will first compute a grounded representation of that message's meaning; to
translate, we find a natural-language message whose meaning is most similar. The
key question is then what form this grounded meaning representation should take.
The existing literature suggests two broad approaches:

\paragraph{Semantic representation} The meaning of a message $\msg_a$ is
given by its denotations: that is, by the set of world states of which
$\msg_a$ may be felicitously predicated, given the existing context available to
a listener. In probabilistic terms, this says that the meaning of a message
$\msg_a$ is represented by the distribution $p(\obs_a | \msg_a, \obs_b)$ it
induces over speaker states. Examples of this approach include
\newcite{Guerin01Denotational} and \newcite{Pasupat16Denotations}.

\paragraph{Pragmatic representation} The meaning of a message $\msg_a$ is
given by the behavior it induces in a listener. In probabilistic terms, this
says that the meaning of a message $\msg_a$ is represented by the distribution
$p(\act_b | \msg_a, \obs_b)$ it induces over actions given the listener's
observation $\obs_b$. Examples of this approach include \newcite{Vogel13Grice}
and \newcite{Gauthier16GoalDriven}.

\paragraph{} These two approaches can give rise to rather different behaviors. Consider the following
example:
\begin{center}
  \vspace{-.3em}
  \includegraphics[width=0.5\columnwidth,clip,trim=0 5.5in 6.5in 0]{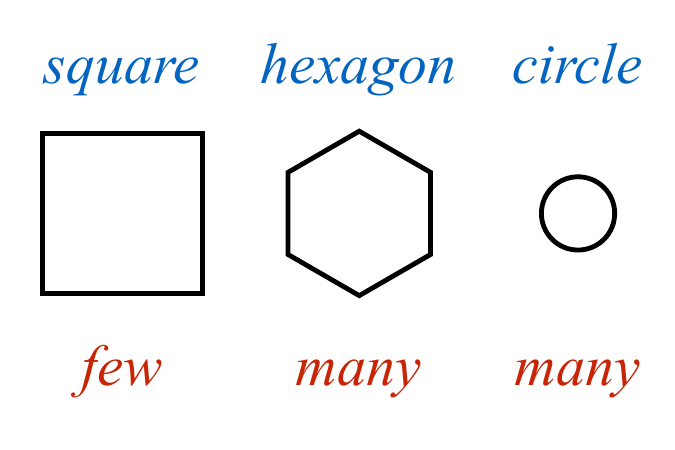}
  \vspace{-.6em}
\end{center}
The top language (in blue) has a unique name for every kind of shape, while the
bottom language (in red) only distinguishes between shapes with few sides and
shapes with many sides. Now imagine a simple reference game with the following
form: player $a$ is covertly assigned one of these three shapes as a reference
target, and communicates that reference to $b$; $b$ must then pull a lever
labeled \texttt{large} or \texttt{small} depending on the size of the target
shape. Blue language speakers can achieve perfect success at this game,
while red language speakers can succeed at best two out of three times.

How should we translate the blue word \emph{hexagon} into the red language? The
semantic approach suggests that we should translate \emph{hexagon} as
\emph{many}: while \emph{many} does not uniquely identify the hexagon, it
produces a distribution over shapes that is closest to the truth. The pragmatic
approach instead suggests that we should translate \emph{hexagon} as \emph{few},
as this is the only message that guarantees that the listener will pull the
correct lever \texttt{large}. So in order to produce a correct listener action,
the translator might have to ``lie'' and produce a maximally inaccurate listener
belief.

If we were exclusively concerned with building a translation layer that allowed
humans and DCP agents to interoperate as effectively as possible, it would 
be natural to adopt a pragmatic representation strategy. But our goals here are
broader: we also want to facilitate \emph{understanding}, and specifically to
help users of learned systems form true beliefs about the systems'
computational processes and representational abstractions. The example above
demonstrates that ``pragmatically'' optimizing directly for task performance can
sometimes lead to translations that produce inaccurate beliefs. 

We instead build our approach around semantic representations of meaning. By
preserving semantics, we allow listeners to reason accurately about the content
and interpretation of messages.  We might worry that by adopting a
semantics-first view, we have given up all guarantees of effective
interoperation between humans and agents using a translation layer.
Fortunately, this is not so: as we will see in \autoref{sec:math}, it is
possible to show that players communicating via a semantic translator perform
only boundedly worse (and sometimes better!) than pairs of players with a common
language.

\section{Translation models}
\label{sec:models}

In this section, we build on the intuition that messages should be translated
via their semantics to define a concrete translation model---a procedure
for constructing a natural language $\leftrightarrow$ neuralese dictionary given 
agent and human interactions.

We understand the meaning of a message 
$\msg_a$ to be represented by the distribution $p(\obs_a|\msg_a, \obs_b)$ it induces 
over speaker states given listener context.
We can formalize this by defining the belief distribution $\belief$ for a message $\msg$ and context 
$\obs_b$ as: \\[-0.3em]

\noindent\scalebox{0.95}{
\[
  \belief(\msg_a, \obs_b) = p(\obs_a | \msg_a, \obs_b) = \frac{p(\msg_a |
  \obs_a) p(\obs_a, \obs_b)}{ \sum_{\obs_a'} p(\msg_a | \obs_a') p(\obs_a', \obs_b)}
  \ .
\]
}

\noindent Here we have modeled the listener as performing a single step of Bayesian
inference, using the listener state 
and the message generation
model 
(by assumption shared between players) to compute the
posterior 
over speaker states. While in general
neither humans nor DCP agents compute explicit representations of this
posterior, past work has found that both humans and suitably-trained neural
networks can be modeled as Bayesian reasoners
\citep{Frank09PragmaticExperiments,Paige16Inference}.

This provides a context-specific representation of belief, but
for messages $z$ and $z'$ to have the same semantics, they
must induce the same belief over \emph{all} contexts in which
they occur. 
In our probabilistic formulation, this introduces an outer expectation over
contexts, providing a final measure $q$ of the
quality of a translation from $z$ to $z'$:

{
  \setlength{\mathindent}{4pt}
\begin{align}
  &q(\msg, \msg') = \expect\big[\kl(\belief(z, X_b)\ ||\ \belief(z', X_b))\ |\ \msg, \msg'\big] \nonumber \\
  &= \sum_{\obs_a, \obs_b} p(\obs_a, \obs_b | \msg, \msg') \nonumber
  \kl(\belief(\msg, \obs_b)\ ||\ \belief(\msg', \obs_b)) \nonumber \\
  &\propto \frac{1}{p(z')} \sum_{\obs_a, \obs_b} p(\obs_a, \obs_b) \cdot p(\msg
  | \obs_a) \cdot p(\msg' | \obs_a) \nonumber \\[-.9em]
  &\qquad\qquad\quad~~~~ \cdot \kl(\belief(\msg, \obs_b)\ ||\ \belief(\msg', \obs_b))\ ;
	\label{eq:q}
\intertext{recalling that in this setting}
  &\kl(\belief\ ||\ \belief') = \sum_{\obs_a} p(\obs_a | \msg, \obs_b) \log \frac{p(\obs_a
  | \msg, \obs_b)}{p(\obs_a | \msg', \obs_b)}
  \nonumber
\end{align}
}

\noindent which is zero when the messages $z$ and $z'$ give rise to identical belief
distributions and increases as they grow more dissimilar.
To translate, we would like to compute $\tr(\msg_r) = \argmin_{\msg_h} q(\msg_r, \msg_h)$ and
$\tr(\msg_h) = \argmin_{\msg_r} q(\msg_h, \msg_r)$.
Intuitively, \autoref{eq:q} says that we will measure the quality of a proposed
translation $\msg \mapsto \msg'$ by asking the following question: in contexts where $\msg$
is likely to be used, how frequently does $\msg'$ induce the same belief about
speaker states as $\msg$?

While this translation criterion directly encodes the semantic notion
of meaning described in \autoref{sec:philosophy}, it is doubly intractable: 
the KL divergence and outer expectation involve a sum over
all observations $\obs_a$ and $\obs_b$ respectively; these sums are not in
general possible to compute efficiently. To avoid this, we approximate
\autoref{eq:q} by sampling. We draw a collection of samples $(\obs_a, \obs_b)$
from the prior over world states, and then generate for each sample a sequence
of distractors $(\obs_a', \obs_b)$ from $p(\obs_a' | \obs_b)$ (we assume
access to both of these distributions from the problem representation). The KL
term in \autoref{eq:q} is computed over each true sample and its distractors,
which are then normalized and averaged to compute the final score.

\begin{algorithm}[t]
  \begin{algorithmic}
    \State
    \State \textbf{given}: a phrase inventory $L$
    \Function{translate}{$z$}
      \State \Return $\argmin_{z' \in L} \hat{q}(z, z')$
    \EndFunction
    \State \vspace{-1em}
    \Function{$\hat{q}$}{$z, z'$}
    \State \textit{// sample contexts and distractors}
    \State $x_{ai}, x_{bi} \sim p(X_a, X_b) \textrm{ for $i=1..n$}$
    \State $x_{ai}' \sim p(X_a | x_{bi})$
    \State \textit{// compute context weights}
    \State $\tilde{w}_i \gets p(z | x_{ai}) \cdot p(z' | x_{ai})$
    \State $w_i \gets \tilde{w}_i / \sum_j \tilde{w}_j$
    \State \textit{// compute divergences}
    \State $ k_i \gets  \sum_{x \in \{x_{ai}, x_{ai}'\}} p(x|z, x_{bi}) \log
    \frac{p(x|z, x_{bi})}{p(x|z', x_{bi})}$
    \State \Return $\sum_i w_i k_i$
    \EndFunction 
    \State 
    \vspace{-1em}
  \end{algorithmic}
  \caption{Translating messages}
  \label{alg:translation}
\end{algorithm}

Sampling accounts for the outer $p(\obs_a, \obs_b)$ in \autoref{eq:q}.
One of the two remaining quantities has the form $p(\obs_a|\msg,\obs_b)$. In the
case of neuralese, can be obtained via Bayes' rule from the agent policy
$\pi_r$. For natural language, we use transcripts of human interactions to fit a
model that maps from frequent utterances to a distribution over world states as
discussed in \autoref{sec:formulation}. The last quantity is a $p(z')$,
the prior probability of the candidate translation; this is approximated as
uniform. The full translation procedure is given in
\autoref{alg:translation}.

\section{Belief and behavior}
\label{sec:math}

The translation criterion in the previous section makes no reference to listener
actions at all. The shapes example in \autoref{sec:philosophy} shows that some model performance might be
lost under translation.  It is thus reasonable to ask whether this translation model
of \autoref{sec:models} can make any guarantees about the effect of translation
on behavior. In this section we explore the relationship between
belief-preserving translations and the behaviors they produce, by examining the
effect of belief accuracy and strategy mismatch on the reward obtained by
cooperating agents.

\begin{figure}
\vspace{-.5em}
  \centering
  \includegraphics[width=0.8\columnwidth,clip,trim=0in 5in 5in 0in]{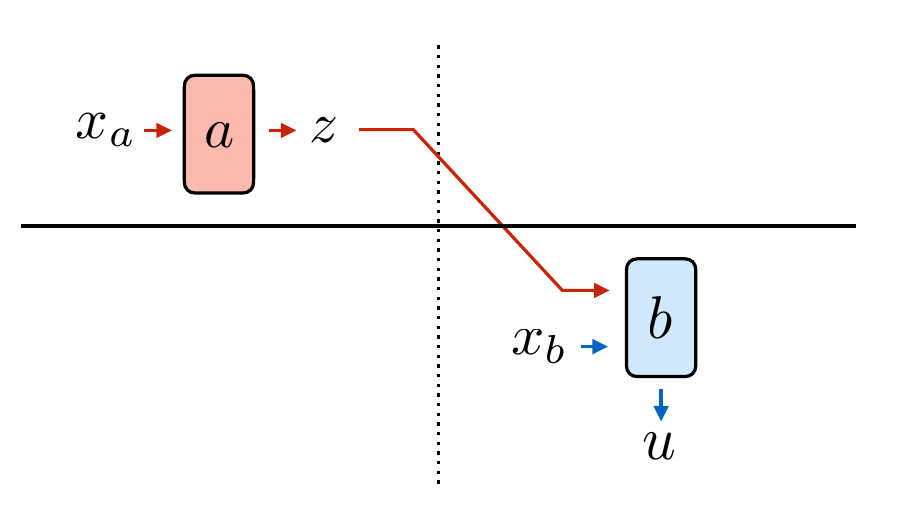} \\
  \vspace{-1.5em}
  \caption{
    Simplified game representation used for analysis in \autoref{sec:math}. 
    A speaker agent sends a message to a listener agent,
    which takes a single action and receives a reward.
  }
  \label{fig:simplegame}
  \vspace{-1.2em}
\end{figure}

To facilitate this analysis, we consider a simplified family of communication
games with the structure depicted in \autoref{fig:simplegame}. These games can
be viewed as a subset of the family depicted in
\autoref{fig:model}; and consist of two steps: a listener makes an observation
$\obs_a$ and sends a single message $\msg$ to a speaker, which makes its own
observation $\obs_b$, takes a single action $u$, and receives a reward. 
We emphasize that the results in this section concern the theoretical properties of idealized games, and are presented to provide intuition about high-level properties of our approach. \autoref{sec:results} investigates empirical behavior of this approach
on real-world tasks where these ideal conditions do not hold.

Our first result is that translations that minimize semantic dissimilarity $q$
cause the listener to take near-optimal actions:\footnote{Proof is
  provided in \autoref{sec:proofs}.}

\pagebreak

\strut
\vspace{-1em}
\hrule

\begin{prop} \label{prop:rational} \strut\\
\textit{Semantic translations reward rational listeners}.\linebreak
  Define a \emph{rational listener} as one that chooses the best action in
  expectation over the speaker's state:
  \[ U(\msg, \obs_b) = \argmax_u \sum_{\obs_a} p(\obs_a | \obs_b, \msg) r(\obs_a, \obs_b, u) \]
  for a reward function $r \in [0, 1]$ that depends only on the two observations and the
  action.%
\footnote{This notion of rationality is a fairly weak one: it
  permits many suboptimal communication strategies, and requires only
  that the listener do as well as possible given a fixed speaker---a
  first-order optimality criterion likely to be satisfied by any
  richly-parameterized model trained via gradient descent.}
  Now let $a$ be a speaker of a language $r$, $b$ be a listener of the same
  language $r$, and $b'$ be a listener of a different language $h$.  Suppose
  that we wish for $a$ and $b'$ to interact via the translator $\tr :
  \msg_r \mapsto \msg_h$ (so that $a$ produces a message $z_r$, and $b'$ takes an
  action $U(\msg_h = \tr(\msg_r), \obs_{b'})$). If $\tr$ respects the semantics of $z_r$, then
  the bilingual pair $a$ and $b'$ achieves only boundedly worse reward than the
  monolingual pair $a$ and $b$.
  Specifically, if $q(\msg_r, \msg_h) \leq D$, then 
  \begin{align}
  &\expect r(X_a, X_b, U(\tr(Z)) \nonumber\\
  &\qquad\geq \expect r(X_a, X_b, U(Z)) - \sqrt{2D} 
  \end{align}
\end{prop}

\vspace{-.5em}
\hrule
\vspace{1em}

So as discussed in \autoref{sec:philosophy}, even by committing to a
semantic approach to meaning representation, we have still succeeded in
(approximately) capturing the nice properties of the pragmatic approach.

\autoref{sec:philosophy} examined the consequences of a mismatch between the set of
primitives available in two languages. In general we would like some measure of
our approach's robustness to the lack of an exact correspondence between two
languages. In the case of humans in particular we expect that a variety of
different strategies will be employed, many of which will not correspond to the
behavior of the learned agent. It is natural to want some assurance that we can
identify the DCP's strategy as long as \emph{some} human strategy mirrors it. 
Our second observation is that it is possible to exactly recover a translation of a DCP
strategy from a mixture of humans playing different strategies:

\vspace{1em}
\hrule

\begin{prop} \label{prop:recovery} \strut \\
\SetTracking{encoding=*}{-30}\lsstyle
\textit{Semantic translations find hidden correspondences}. \linebreak
\SetTracking{encoding=*}{0}\lsstyle
Consider a fixed robot policy $\pi_r$ and a set of human policies
  $\{\pi_{h1}, \pi_{h2}, \dots\}$ (recalling from \autoref{sec:formulation} that each $\pi$ is
  defined by distributions $p(z|x_a)$ and $p(u|z,x_b)$). Suppose further that
  the messages employed by these human strategies are \emph{disjoint}; that is,
  if $p_{hi}(z|x_a) > 0$, then $p_{hj}(z|x_a) = 0$ for all $j \neq i$. Now
  suppose that all $q(\msg_r, \msg_h) = 0$ for all messages in the support
  of some $p_{hi}(z|x_a)$ and $> 0$ for all $j \neq i$. Then every message $z_r$ is 
  translated into a message produced by $\pi_{hi}$, and messages from other strategies
  are ignored.

\vspace{0.5em}
\hrule
\end{prop} 
\vspace{1em}

This observation follows immediately from the definition of $q(\msg_r, \msg_h)$, but
demonstrates one of the key distinctions between our approach and a conventional machine
translation criterion. Maximizing $p(\msg_h | \msg_r)$ will produce the natural language
message most often produced in contexts where $\msg_r$ is observed, regardless of whether
that message is useful or informative. By contrast, minimizing $q(\msg_h, \msg_r)$ will 
find the $\msg_h$ that corresponds most closely to $\msg_r$ even when $\msg_h$ is rarely 
used.

\begin{figure}
  \centering
  \includegraphics[width=\columnwidth,clip,trim=.1in 2in 5.55in .1in]{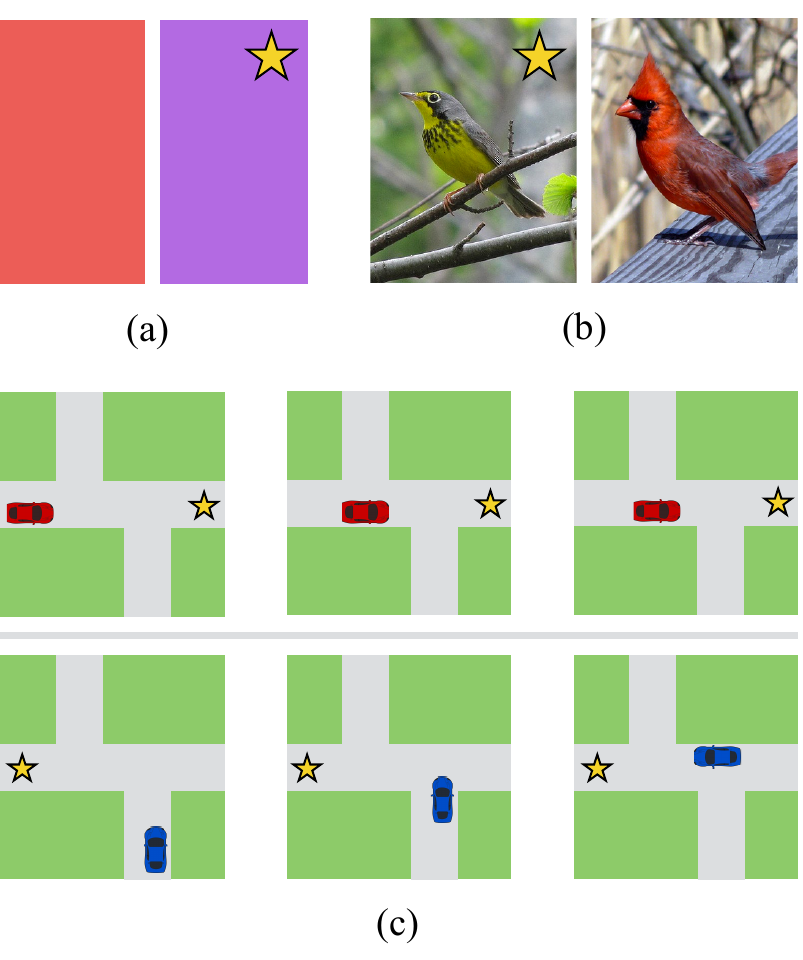}
  \vspace{-1.8em}
  \caption{Tasks used to evaluate the translation model. (a--b) Reference games:
    both players observe a pair of reference candidates (colors or images);
    Player $a$ is assigned a target (marked with a star), which player $b$ must
    guess based on a message from $a$. (c) Driving game: each car attempts to
    navigate to its goal (marked with a star). The cars cannot see each other,
    and must communicate to avoid a collision.
  }
	\label{fig:tasks}
    \vspace{-1em}
\end{figure}

The disjointness condition, while seemingly quite strong, in fact arises
naturally in many circumstances---for example, players in the driving game
reporting their spatial locations in absolute vs.\ relative coordinates, or
speakers in a color reference game (\autoref{fig:tasks}) discriminating based on
lightness vs.\ hue. It is also possible to relax the above condition to require
that strategies be only \emph{locally} disjoint (i.e.\ with the disjointness
condition holding for each fixed $\obs_a$), in which case overlapping human
strategies are allowed, and the recovered robot strategy is a context-weighted
mixture of these.

\section{Evaluation}

\subsection{Tasks}

In the remainder of the paper, we evaluate the empirical behavior of our
approach to translation.
Our evaluation considers two kinds of tasks: reference games and navigation
games. In a reference game (e.g.\ \autoref{fig:tasks}a), both players observe a
pair of candidate referents. A speaker is assigned
a target referent; it must communicate this target to a listener, who then
performs a choice action corresponding to its belief about the true target.
In this paper we consider two variants on the reference game: a simple
color-naming task, and a more complex task involving natural images of birds.
For examples of human communication strategies for these tasks, we obtain
the XKCD color dataset \cite{McMahan15Colors,Monroe16Colors} and the
Caltech--UCSD Birds dataset \cite{Welinder10Birds} with accompanying natural
language descriptions \cite{Reed16Birds}. We use standard train / validation /
test splits for both of these datasets.

The final task we consider is the driving task (\autoref{fig:tasks}c) first
discussed in the introduction. In this task, two cars, invisible to each
other, must each navigate between randomly assigned start and goal positions
without colliding. This task takes a number of steps to complete, and
potentially involves a much broader range of communication strategies. To
obtain human annotations for this task, we recorded both actions and messages
generated by pairs of human Amazon Mechanical Turk workers playing the driving
game with each other. We collected close to 400 games, with a total of more
than 2000 messages exchanged, from which we held out 100 game traces as a test
set.

\subsection{Metrics}

A mechanism for
understanding the behavior of a learned model should allow a human user both to
correctly infer its beliefs and to successfully interoperate with it;
we accordingly report results of both ``belief'' and ``behavior'' evaluations.

To support easy reproduction and comparison
(and in keeping with standard practice in machine translation), we focus on developing automatic
measures of system performance. We use the available training data to develop
simulated models of human decisions; by first showing that these
models track well with human judgments, we can be confident that their use in
evaluations will correlate with human understanding. We employ the following two
metrics:

\paragraph{Belief evaluation} This evaluation focuses on the denotational
perspective in semantics that motivated the initial development of our model.
We have successfully understood the semantics of a
message $\msg_r$ if, after translating $\msg_r \mapsto \msg_h$, a
human listener can form a correct belief about the state in which $\msg_r$ was
produced. We construct a simple state-guessing game where the
listener is presented with a translated message and two state observations, and
must guess which state the speaker was in when the message was emitted. 

When translating from natural language to neuralese, we use the learned agent
model to directly guess the hidden state.  For neuralese to natural language we
must first construct a ``model human listener'' to map from strings back to
state representations; we do this by using the training data to fit a simple
regression model that scores (state, sentence) pairs using a bag-of-words
sentence representation. We find that our ``model human'' matches the judgments
of real humans 83\% of the time on the colors task, 77\% of the time on the
birds task, and 77\% of the time on the driving task.  This gives us confidence
that the model human gives a reasonably accurate proxy for human interpretation.

\paragraph{Behavior evaluation} This evaluation focuses on the cooperative
aspects of interpretability: we measure the extent to which learned models are
able to interoperate with each other by way of a translation layer.
In the case of reference games, the goal of this semantic evaluation is
identical to the goal of the game itself (to identify the hidden state of the
speaker), so we perform this additional pragmatic evaluation only for the
driving game. We found that the most reliable way to make use of human game
traces was to construct a \emph{speaker-only} model human. The evaluation
selects a full game trace from a human player, and replays both the human's
actions and messages exactly (disregarding any incoming messages); the
evaluation measures the quality of the natural-language-to-neuralese translator,
and the extent to which the learned agent model can accommodate a (real) human
given translations of the human's messages.

\paragraph{Baselines} We compare our approach to two baselines: a \emph{random}
baseline that chooses a translation of each input uniformly from messages
observed during training, and a \emph{direct} baseline that directly maximizes
$p(\msg' | \msg)$ (by analogy to a conventional machine translation system).
This is accomplished by sampling from a DCP speaker in training states labeled
with natural language strings.

\section{Results}
\label{sec:results}

\begin{table}[t]
\vspace{-.5em}
  \centering
  \footnotesize
  \raisebox{-1.2em}{(a)}\hspace{1em}
  \begin{tabular}{cc|c|c|l}
    & \multicolumn{1}{c}{} & \multicolumn{2}{c}{as speaker} \\
    & \multicolumn{1}{c}{} & \multicolumn{1}{c}{R} & \multicolumn{1}{c}{H} \\
    \cline{3-4}
    \multirow{6}{*}{\rotatebox[origin=c]{90}{as listener}} 
    & \multirow{3}{*}{R} & \multirow{3}{*}{
      1.00 
    } & 
      0.50 
      & random
    \\
    \cline{4-4}
    & & &
      0.70 
      & direct
    \\
    \cline{4-4}
    & & & 
      \bf 0.73 
      & belief (ours)
    \\
    \cline{3-4}
    & \multirow{3}{*}{H*} & 
      0.50 
    & \multirow{3}{*}{
      0.83 
    } \\
    \cline{3-3}
    & & 
      0.72 
    & \\
    \cline{3-3}
    & &
      \bf 0.86 
    & \\
    \cline{3-4}
  \end{tabular}

  \vspace{1em}

  \raisebox{-1.2em}{(b)}\hspace{1em}
  \begin{tabular}{cc|c|c|l}
    & \multicolumn{1}{c}{} & \multicolumn{2}{c}{as speaker} \\
    & \multicolumn{1}{c}{} & \multicolumn{1}{c}{R} & \multicolumn{1}{c}{H} \\
    \cline{3-4}
    \multirow{6}{*}{\rotatebox[origin=c]{90}{as listener}} 
    & \multirow{3}{*}{R} & \multirow{3}{*}{
      0.95 
    } & 
      0.50 
      & random
    \\
    \cline{4-4}
    & & & 
      0.55 
      & direct
    \\
    \cline{4-4}
    & & & 
      \bf 0.60 
      & belief (ours)
    \\
    \cline{3-4}
    & \multirow{3}{*}{H*} & 
      0.50 
    & \multirow{3}{*}{
      0.77 
    } \\
    \cline{3-3}
    & & 
      0.57 
    & \\
    \cline{3-3}
    & & 
      \bf 0.75 
    & \\
    \cline{3-4}
  \end{tabular}

  \caption{
    Evaluation results for reference games. (a) The colors task. (b) The birds
    task. Whether the model human is in a listener or speaker role, translation
    based on belief matching outperforms both random and machine translation
    baselines. 
  }
  \label{tab:reference}
\end{table}

\begin{figure}[b]
  \centering
  \includegraphics[width=\columnwidth,clip,trim=.1in 2.8in 1.7in .1in]{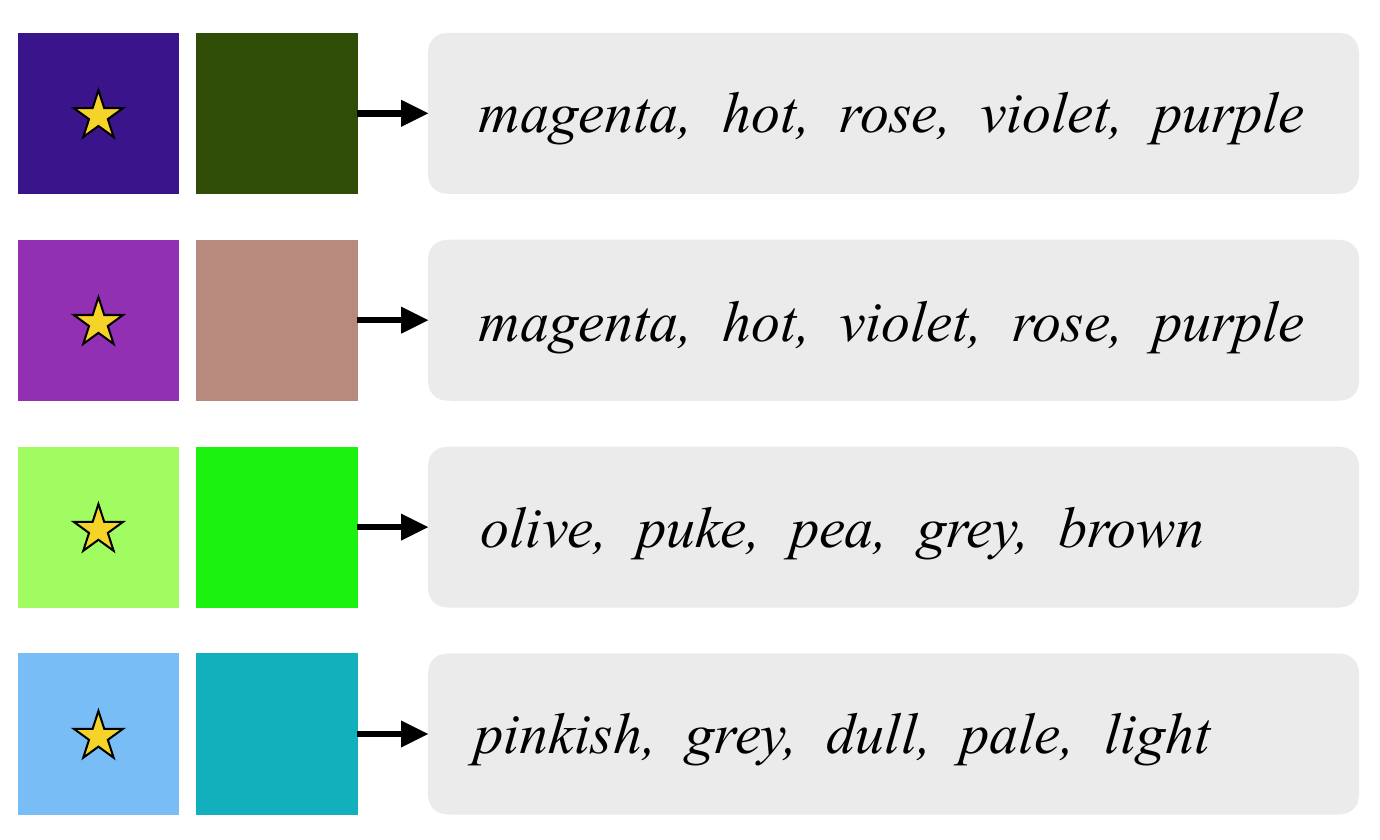}
  \caption{Best-scoring translations generated for color task.}
  \label{fig:color-examples}
\end{figure}

In all below, ``R'' indicates a DCP agent, ``H'' indicates a real human,
and ``H*'' indicates a model human player.

\paragraph{Reference games}

Results for the two reference games are shown in \autoref{tab:reference}. 
The end-to-end trained model achieves nearly perfect accuracy in
both cases, while a model trained to communicate in natural language achieves
somewhat lower performance. Regardless of whether the speaker is a DCP and the
listener a model human or vice-versa, translation based on the belief-matching
criterion in \autoref{sec:models} achieves the best performance; indeed, when
translating neuralese color names to natural language, the listener is able to
achieve a slightly higher score than it is natively. This suggests that the 
automated
agent has discovered a more effective strategy than the one demonstrated by
humans in the dataset, and that the effectiveness of this strategy is preserved
by translation.  Example translations from the reference games are depicted in
\autoref{fig:bird-examples} and \autoref{fig:color-examples}. 

\begin{table}[t]
  \centering
  \footnotesize
  \begin{tabular}{cc|c|c|l}
    & \multicolumn{1}{c}{} & \multicolumn{2}{c}{as speaker} \\
    & \multicolumn{1}{c}{} & \multicolumn{1}{c}{R} & \multicolumn{1}{c}{H} \\
    \cline{3-4}
    \multirow{6}{*}{\rotatebox[origin=c]{90}{as listener}} 
    & \multirow{3}{*}{R} & \multirow{3}{*}{
      0.85 
    } & 
      0.50 
      & random
    \\
    \cline{4-4}
    & & & 
      0.45 
      & direct
    \\
    \cline{4-4}
    & & & 
      \bf 0.61 
      & belief (ours)
    \\
    \cline{3-4}
    & \multirow{3}{*}{H*} & 
      0.5 
    & \multirow{3}{*}{
      0.77 
    } \\
    \cline{3-3}
    & & 
      0.45 
    & \\
    \cline{3-3}
    & & 
      \bf 0.57 
    & \\
    \cline{3-4}
  \end{tabular}
  \caption{
    Belief evaluation results for the driving game. Driving states are
    challenging to identify based on messages alone (as evidenced by the
    comparatively low scores obtained by single-language pairs) .
    Translation based on belief
    achieves the best overall performance in both directions.
  }
  \label{tab:driving-sem}
\end{table}

\begin{table}[t]
  \centering
  \footnotesize
  \begin{tabular}{|c|c|c|l}
    \cline{1-3}
    R / R & H / H & R / H \\
    \cline{1-3}
    \multirow{3}{*}{
      1.93 / 0.71 
    } & \multirow{3}{*}{
      --- / 0.77 
    } &
      1.35 / 0.64 
      & random \\ \cline{3-3} & &
      1.49 / \bf 0.67 
      & direct \\ \cline{3-3} & &
      \bf 1.54 / 0.67 
      & belief (ours) \\
    \cline{1-3}
  \end{tabular}
  \caption{
    Behavior evaluation results for the driving game. 
    Scores are presented in the form ``reward / completion 		
    rate''.
    While less accurate than
    either humans or DCPs with a shared language, the models that employ a
    translation layer obtain higher reward and a greater overall success
    rate than baselines. 
  }
  \label{tab:driving-prag}
  \vspace{-1em}
\end{table}

\paragraph{Driving game}

\begin{figure}
  \vspace{-.5em}
  \centering
  \includegraphics[width=.85\columnwidth,clip,trim=0 .2in 3.5in 0]{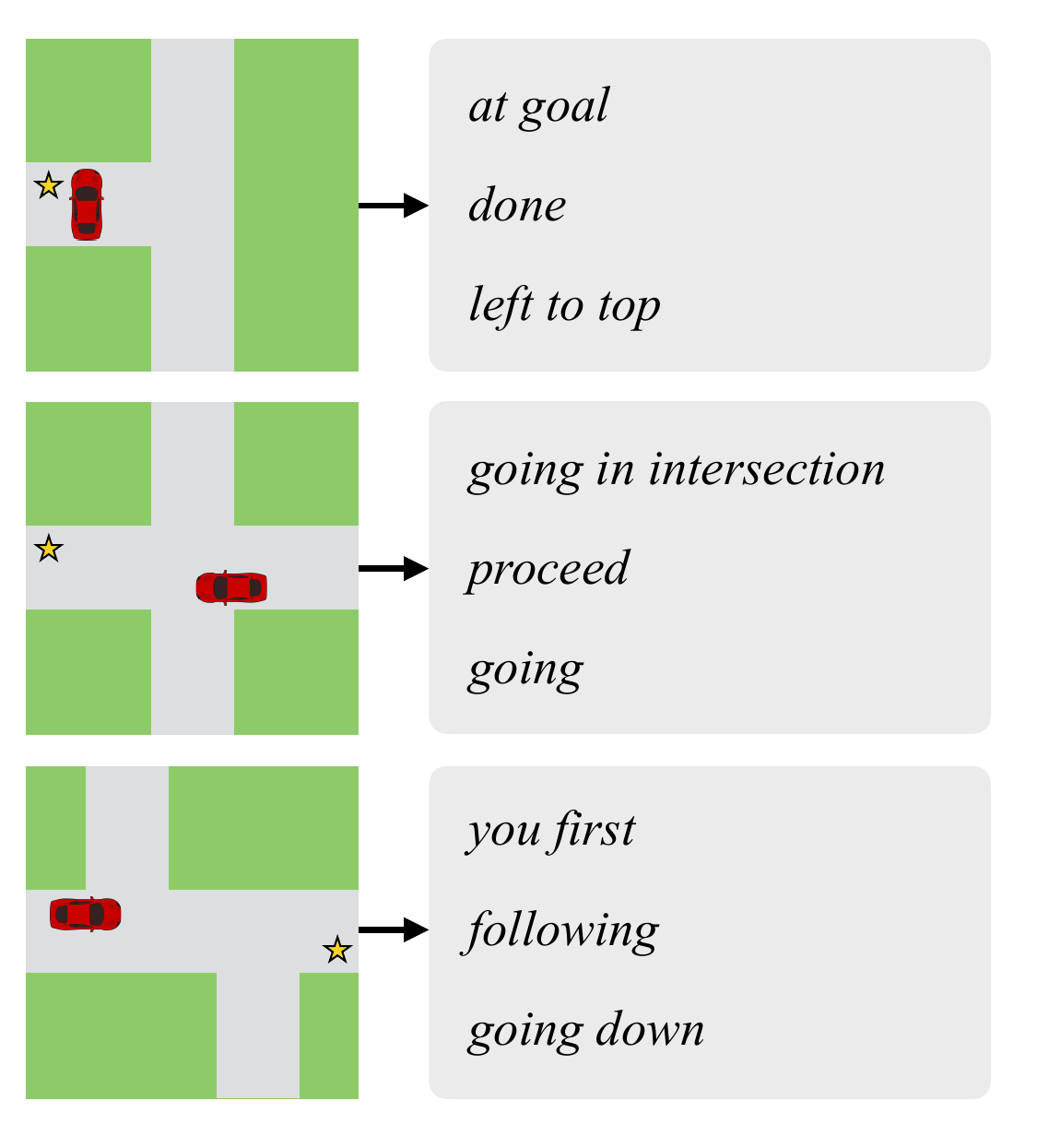}
  \vspace{-1em}
  \caption{Best-scoring translations generated for driving task generated from
  the given speaker state.}
  \label{fig:drive-examples}
  \vspace{-1em}
\end{figure}

Behavior evaluation of the driving game is shown in \autoref{tab:driving-prag},
and belief evaluation is shown in \autoref{tab:driving-sem}. Translation of
messages in the driving game is considerably more challenging than in the
reference games, and scores are uniformly lower; however, a clear benefit from
the belief-matching model is still visible. Belief matching leads to higher
scores on the belief evaluation in both directions, and allows agents to
obtain a higher reward on average (though task completion rates remain roughly
the same across all agents).
Some example translations of driving game messages are shown in
\autoref{fig:drive-examples}.

\section{Conclusion}

We have investigated the problem of interpreting message vectors from deep
networks by translating them. After introducing a translation criterion based on
matching listener beliefs about speaker states, we presented both theoretical
and empirical evidence that this criterion outperforms a conventional
machine translation approach at recovering the content of message vectors
and facilitating collaboration between humans and learned agents.

While our evaluation has focused on understanding the behavior of deep
communicating policies, the framework proposed in this paper could be much more generally
applied. Any encoder--decoder model \cite{Sutskever14NeuralSeq} can be thought
of as a kind of communication game played between the encoder and the decoder,
so we can analogously imagine computing and translating ``beliefs'' induced by
the encoding to explain what features of the input are being transmitted. The
current work has focused on learning a purely categorical model of the
translation process, supported by an unstructured inventory of translation
candidates, and future work could explore the \emph{compositional} structure of
messages, and attempt to synthesize novel natural language or neuralese messages
from scratch. 
More broadly, the work here shows that the denotational perspective from formal
semantics provides a framework for precisely framing the demands of
interpretable machine learning \cite{Wilson16Interpretable}, and particularly
for ensuring that human users without prior exposure to a learned model are able
to interoperate with it, predict its behavior, and diagnose its errors. 

\section*{Acknowledgments}
JA is supported by a Facebook Graduate Fellowship and a Berkeley AI / Huawei
Fellowship. We are grateful to Lisa Anne Hendricks for assistance with the
Caltech--UCSD Birds dataset, and to Liang Huang and Sebastian Schuster for useful
feedback.

\bibliography{jacob}
\bibliographystyle{acl_natbib}

\newpage
\onecolumn

\appendix

\section{Proofs}
\label{sec:proofs}

\paragraph{Proof of \autoref{prop:rational}}

We know that
\[ U(\msg, \obs_b) := \argmax_u \sum_{\obs_a} p(\obs_a|\obs_b, \msg) r(\obs_a,
\obs_b, \msg) \]
and that for all translations $(z, z'=t(r))$
\begin{align*}
  &D \geq \sum_{\obs_b} p(\obs_b | z, z') \kl(\belief(z, \obs_b)\ ||\
  \belief(z', \obs_b))\ . \\
 \intertext{Applying Pinsker's inequality:}
  &\geq 2 \sum_{\obs_b} p(\obs_b | z, z') \delta(\belief(z, \obs_b), \belief(z', \obs_b))^2 \\
  \intertext{and Jensen's inequality:}
  &\geq 2\bigg(\sum_{\obs_b} p(\obs_b | z, z') \delta(\belief(z, \obs_b), \belief(z', \obs_b)))\bigg)^2 \\
 \intertext{so}
 &\sqrt{D/2} \geq \sum_{\obs_b} p(\obs_b | z, z') \delta(\belief(z, \obs_b),
 \belief(z',\obs_b)) \ .
\end{align*}
The next step relies on the following well-known property of the total variation distance: for distributions $p$ and $q$ and a function $f$ bounded by $[0, 1]$,
\[ |\expect_p f(x) - \expect_q f(x)| \leq \delta(p, q) \ . \tag{*} \]
For convenience we will write 
\[ \delta := \delta(\belief(z, \obs_b), \belief(z', \obs_b))\ . \]
A listener using the speaker's language expects a reward of
\begin{align*}
	&\sum_{\obs_b} p(\obs_b) \sum_{\obs_a} p(\obs_a | \obs_b, z) r(\obs_a, \obs_b,
  U(z, x_b)) \\
    &\leq \sum_{\obs_b} p(\obs_b) \bigg(\sum_{\obs_a} p(\obs_a | \obs_b, z')
    r(\obs_a, \obs_b, U(z, x_b)) + \delta\bigg)
    \intertext{via (*). From the assumption of player rationality:}
    &\leq \sum_{\obs_b} p(\obs_b) \bigg(\sum_{\obs_a} p(\obs_a | \obs_b, z')
    r(\obs_a, \obs_b, U(\textcolor{red}{z'}, x_b)) + \delta\bigg) \\ 
    \intertext{using (*) again:}
    &\leq \sum_{\obs_b} p(\obs_b) \bigg(\sum_{\obs_a} p(\obs_a | \obs_b,
    \textcolor{red}{z}) r(\obs_a, \obs_b, U(z', x_b)) + 2\delta\bigg) \\
    &\leq \sum_{\obs_a, \obs_b}p(\obs_a, \obs_b | z)r(x_a, x_b, U(z', x_b)) +
    \sqrt{2D} \ .
\end{align*}
So the true reward achieved by a $z'$-speaker receiving a translated code is
only additively worse than
the native $z$-speaker reward:
\[ \bigg(\sum_{\obs_a, \obs_b} p(x_a, x_b | z) r(x_a, x_b, U(z, x_b))\bigg) -
\sqrt{2D} \hfill \qed \]

\newpage
\twocolumn

\section{Implementation details}
\label{sec:impl}

\subsection{Agents}

Learned agents have the following form:

\noindent \includegraphics[width=2in,clip,trim=0in 5in 4.7in 0.1in]{figs/cell}

\noindent where $h$ is a hidden state, $z$ is a message from the other agent, $u$ is a
distribution over actions, and $x$ is an observation of the world. A single
hidden layer with 256 units and a $\tanh$ nonlinearity is used for the MLP. The
GRU hidden state is also of size 256, and the message vector is of size 64.

Agents are trained via interaction with the world as in
\newcite{Hausknecht15DRQN} using the \textsc{adam} optimizer \cite{Kingma14Adam}
and a discount factor of 0.9. The step size was chosen as $0.003$ for reference
games and $0.0003$ for the driving game.
An $\epsilon$-greedy exploration strategy is
employed, with the exploration parameter for timestep $t$ given by:\\[-0.5em]
\[
  \epsilon = \max \begin{cases}
    (1000 - t) / 1000 \\
    (5000 - t) / 50000 \\
    0
  \end{cases}
\]

As in \newcite{Foerster16Communication}, we found it useful to add noise to the
communication channel: in this case, isotropic Gaussian noise with mean 0 and
standard deviation 0.3. This also helps smooth $p(z|x_a)$
when computing the translation criterion.

\subsection{Representational models}

As discussed in \autoref{sec:models}, the translation criterion is computed
based on the quantity $p(z|x)$. The policy representation above actually
defines a distribution $p(z|x, h)$, additionally involving the agent's hidden
state $h$ from a previous timestep. While in principle it is possible to
eliminate the dependence on $h$ by introducing an additional sampling step into
\autoref{alg:translation}, we found that it simplified inference to simply learn
an additional model of $p(z|x)$ directly. For simplicity, we treat the term
$\log(p(z') / p(z))$ as constant, those these could be more accurately
approximated with a learned density estimator.

This model is trained alongside the learned agent to
imitate its decisions, but does not get to observe the recurrent state, like so:

\noindent \includegraphics[width=1.6in,clip,trim=0in 6.5in 6.5in 0in]{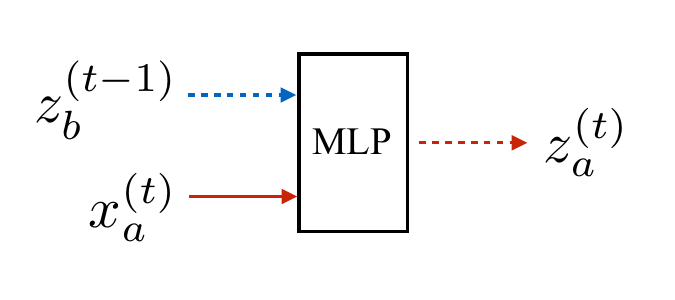}

\noindent Here the multilayer perceptron has a single hidden layer with $\tanh$
nonlinearities and size 128. It is also trained with \textsc{adam} and a step
size of 0.0003.

We use exactly the same model and parameters to implement representations of
$p(z|x)$ for human speakers, but in this case the vector $z$ is taken to be a
distribution over messages in the natural language inventory, and the model is
trained to maximize the likelihood of labeled human traces.

\subsection{Tasks}

\paragraph{Colors}

We use the version of the XKCD dataset prepared by \newcite{McMahan15Colors}.
Here the input feature vector is simply the LAB representation of each color,
and the message inventory taken to be all unigrams that appear at least five times.

\paragraph{Birds}

We use the dataset of \newcite{Welinder10Birds} with natural language
annotations from \newcite{Reed16Birds}. The model's input feature
representations are a final 256-dimensional hidden feature vector from a
compact bilinear pooling model \cite{Gao16CBP} pre-trained for
classification. The message inventory consists of the 50 most frequent
bigrams to appear in natural language descriptions; example human traces are
generated by for every frequent (bigram, image) pair in the dataset.

\paragraph{Driving}

Driving data is collected from pairs of human workers on Mechanical Turk.
Workers received the following description of the task:
\begin{quote}
  Your goal is to drive the red car onto the red square. Be careful! You're
  driving in a thick fog, and there is another car on the road that you cannot
  see. However, you can talk to the other driver to make sure you both reach
  your destinations safely.
\end{quote}
Players were restricted to messages of 1--3 words, and required to send at
least one message per game. Each player was paid \$0.25 per game. 382 games were
collected with 5 different road layouts, each represented as an 8x8 grid
presented to players as in \autoref{fig:drive-examples}. The action space is
discrete: players can move forward, back, turn left, turn right, or wait.
These were divided into a 282-game training set and 100-game test set. The
message inventory consists of all messages sent more than 3 times. Input
features consists of indicators on the agent's current position and orientation,
goal position, and map identity. Data is available for download at \linebreak
\scalebox{0.83}{\url{http://github.com/jacobandreas/neuralese}}.

\end{document}